# Low-rank Tensor Assisted K-space Generative Model for Parallel Imaging Reconstruction

Wei Zhang, Zengwei Xiao, Hui Tao,
Minghui Zhang, Xiaoling Xu, Qiegen Liu, *Senior Member, IEEE*

*Abstract*—Although recent deep learning methods, especially generative models, have shown good performance in fast magnetic resonance imaging, there is still much room for improvement in high-dimensional generation. Considering that internal dimensions in score-based generative models have a critical impact on estimating the gradient of the data distribution, we present a new idea, low-rank tensor assisted k-space generative model (LR-KGM), for parallel imaging reconstruction. This means that we transform original prior information into high-dimensional prior information for learning. More specifically, the multi-channel data is constructed into a large Hankel matrix and the matrix is subsequently folded into tensor for prior learning. In the testing phase, the low-rank rotation strategy is utilized to impose low-rank constraints on tensor output of the generative network. Furthermore, we alternately use traditional generative iterations and low-rank high-dimensional tensor iterations for reconstruction. Experimental comparisons with the state-of-the-arts demonstrated that the proposed LR-KGM method achieved better performance.

*Index Terms*—Parallel imaging reconstruction, generative model, high-dimensional tensor, Hankel matrix.

## I. INTRODUCTION

Magnetic Resonance Imaging (MRI) as a non-invasive, radiation-free, and in-vivo imaging technique, provides significantly better soft-tissue contrast and offers accurate measurements of both anatomical and functional signal. However, a major drawback of MRI is the long acquisition time which will cause non-idealized space resolution, patients discomfort and hindering applications in time-critical diagnose. To alleviate the prolonged acquisition time, researchers have made great efforts. Obviously, Compressed Sensing (CS) [1]-[4] and Parallel Imaging (PI) [5]-[7] have achieved significant progress in fast MRI.

With PI techniques, the sensitivities of multiple receive coils provide additional information for the reconstruction, allowing for successful reconstruction from sparser sampling. PI techniques can be divided into two categories: reconstruction methods in the image domain and methods in the k-space domain, which require the estimation of missing harmonic data before reconstruction. Typical representatives of these two methods are sensitivity encoding (SENSE) [6] and generalized auto-calibrating partially parallel acquisitions (GRAPPA) [7] respectively. Techniques such as SENSE, explicitly require sensitivity information for reconstruction. Theoretically, SENSE leverages the acquired sensitivity information of multiple receiving coils to remove aliasing artifacts. In practice, it is often difficult to accurately and robustly obtain the coil sensitivity information. For GRAPPA, the essential idea is to estimate the missing k-space data points by linearly combining its acquired neighboring points. However, these methods often require additional k-space data, so-called auto-calibration signal (ACS) lines, to estimate GRAPPA kernels.

With the continuous development of PI, structured low-rank matrix methods have also been commonly utilized [8]-[13]. These methods jointly manipulates multi-coil k-space data by organizing acquired data into a single, structured matrix. In recent years, the structured low-rank matrix priors have been shown to achieve better reconstruction results. Specifically, simultaneous auto-calibrating and k-space estimation [14] explore the strong correlation among receiver coils to enable the low-rankness of a block Hankel matrix constructed from the multi-coil k-space data. The low-rank matrix modeling of local k-space neighborhoods (LORAKS) [15] assumes the image of interest has limited support and a slowly varying phase, which would lead to a block-Hankel-like matrix being low-rank. Inspired by k-space interpolation methods, an annihilating filter-based low-rank Hankel matrix approach (ALOHA) [16] is proposed as a general framework for sparsity-driven k-space interpolation method. As a multi-dimensional generalization of matrix computation, tensor computation has been recently employed in MRI due to its capability of handling high-dimensional data [17]-[22]. In many applications, MRI data sets naturally have a higher physical dimension. In these cases, tensors often capture the hidden high-dimensional data pattern better, achieving better reconstruction performance. So far, most of the tensor methods in multi-dimensional MRI have been employed based on a higher-order singular value decomposition (HOSVD). For example, Liu *et al.* [23] constructed multi-channel k-space data from multiple slices into block-wise Hankel tensors and iteratively updates them by boosting the tensor low-rankness using HOSVD. Later, Zhao *et al.* [24] designed a method for multi-slice acquisition in quadrature alternating phase encoding (PE) directions and subsequent joint calibration-free reconstruction.

Inspired by the great success of deep learning [25]-[30], many researchers have found that deep learning-based MRI reconstruction methods have achieved great success and demonstrated significant performance gains. Recent years, unsupervised deep learning methods, especially deep generative models, have shown great potential in learning object distributions. Deep generative models have advantages of alleviating the deficiency of learning flexibility [31]-[39]. For instance, Song *et al.* [37] introduced a new generative model named noise conditional score networks (NCSN), where samples are produced via Langevin dynamics using

This work was supported in part by National Natural Science Foundation of China under 61871206, 62161026. (Corresponding authors: X. Xu and Q. Liu.) (W. Zhang and Z. Xiao are co-first authors.)

W. Zhang, Z. Xiao, H. Tao, M. Zhang, X. Xu, and Q. Liu are with School of Information Engineering, Nanchang University, Nanchang 330031, China. ({zhangwei1, xiaozengwei, taohui}@email.ncu.edu.cn, {zhangminghui, xuxiaoling, liuqiegen}@ncu.edu.cn)

gradients of the data distribution estimated with DSM progressively. Later, Song et al. [38] proposed a score-based diffusion model by looking at the stochastic differential equation (SDE) associated with the inference process. Similarly, Quan et al. [39] presented HGGDP for MRI reconstruction by taking advantage of the denoising score matching. Although these deep generative models have played an important role in speeding up MRI reconstruction, challenges remain.

In recent years, researchers have achieved breakthroughs in many fields by embedding low-rank into generative models [40]-[44]. For instance, Ding et al. [43] designed two-stage generative adversarial networks to enhance the generalizability of semantic dictionary through low-rank embedding for zero-shot learning in field of visual recognition. In statistics and machine learning, to reconstruct a rank-one signal matrix from noisy data, Cocola et al. [44] studied an alternative prior where the low-rank component is in the range of a trained generative network. In our work, by means of incorporating the strengths of low-rank and recent deep generative model, we introduce a low-rank tensor assisted k-space generative model for MRI reconstruction.

In detail, we construct a large Hankel matrix from multi-channel MRI data and the patches extracted from the matrix are constructed as higher-dimensional tensors for learning. At the reconstruction stage, the corresponding data densities in the tensor is estimated by inferring missing information through the matching network. The problem of redundant information in the SDE output tensor is settled by using low-rank rotation strategy and low-rank constraint is imposed on tensor. Finally, k-space data is recovered from the imposed data consistency by minimizing the difference between estimated k-space and acquired data.

The main contributions of this work are summarized as follows:
- ***High-dimensional tensor for generative model.*** The object generated in this paper is high-dimensional tensor constructed from Hankel matrix, rather than native k-space data, which can further increase the number of input channels in the model and the diversity of generation. Specifically, we first construct a large Hankel matrix from multi-channel MRI data. Then, the high-dimensional tensor is obtained by unfolding the Hankel matrix and applied to the diffusion model.
- ***Iterative generation for alternative reconstruction.*** Update the solution by alternating traditional and generative iterations. In detail, the alternations of low-rank high-dimensional tensor iterations and diffusion generation iterations constitute the whole iterative process, which is beneficial to further improve the reconstruction performance.

The manuscript is organized as follows. Relevant background on structured low-rank matrix and score-based diffusion models are described in Section II. Detailed procedure and algorithm of the proposed methods are presented in Section III. Experimental results and specifications about the implementation and experiments are given in Section IV. At last, we conclude our work in Section V.

## II. RELATED WORK

### A. Prior Learning in PI

The data acquisition for multi-coil PI in k-space can be described as follows:

$$y_c = Mk_c + n_c, \quad c = 1, 2, \cdots, C \quad (1)$$

where $y_c$ denotes the partially observed measurement of the $c$-th coil. $M$ is a diagonal matrix whose diagonal elements are either 1 or 0 depending on the sampling mask. $k_c$ stands for the $c$-th coil k-space data while $n_c$ is the noise. To estimate the reconstructed image of $C$ coils, the optimization problem can be expressed as:

$$\min_k \sum_{c=1}^{C} \| M_c k_c - y_c \|_2^2 + \lambda R(k_c) \quad (2)$$

where the first term represents data fidelity. $\lambda$ is a balance parameter that determines the trade-off between the prior information and the data fidelity term. $R(k_c)$ enforces prior information to improve reconstruction performance.

### B. Structured Low-rank Data Matrix

Utilizing the correlations in multi-channel MRI k-space data, we construct the multi-channel data into a data matrix. The columns of Hankel matrix are vectorized blocks selected by sliding a window over the entire data. For sufficiently large images, the matrix will have more columns than rows. Fig. 1. shows a pictorial depiction of building such a matrix with an exemplary 3×3 window. From $N_x \times N_y$ sized data with $N_c$ number of coils, we can generate a data matrix having the size of $w^2 N_c \times (N_x - w + 1)(N_y - w + 1)$ by sliding a $w \times w \times N_c$ window across the entire k-space. Note that due to the nature of the sliding-window operation, the data matrix will have a stacked, block-wise Hankel structure with many of its entries from identical k-space locations being repeated in antidiagonal directions.

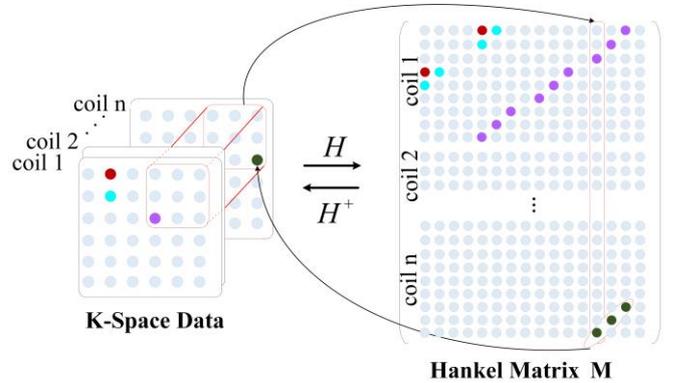

**Fig. 1.** Schematic diagram of the construction of the Hankel matrix.

We define a linear operator $H$ that generates a data matrix **M** from a multi-channel dataset.

$$H: C^{N_x \times N_y \times N_c} \to C^{w^2 N_c \times (N_x - w + 1)(N_y - w + 1)} \quad (3)$$

Then, the reverse operator $H^+$ generates the corresponding k-space dataset from the data matrix **M**.

$$H^+: C^{w^2 N_c \times (N_x - w + 1)(N_y - w + 1)} \to C^{N_x \times N_y \times N_c} \quad (4)$$

where + denotes a pseudo-inverse operator. In other words, the role of $H^+$ is to first enforce the block-wise Hankel structure by averaging the multiple anti-diagonal entries that would have originated from the same k-space locations. Once the data matrix has this structure, $H^+$ stores the averaged values in the appropriate k-space locations (Fig. 1).

### C. Score-based Generative Model with SDE

Score-based generative models which define a forward diffusion process for transforming data to noise and generating data from noise by reversing the forward process have gained a lot of successes in generating realistic and diverse data recently. Recent years, Song *et al.* [38] presented an SDE that transforms a complex data distribution to a known prior distribution by slowly injecting noise and a corresponding reverse-time SDE that transforms the prior distribution back into the data distribution by slowly removing the noise.

More specifically, we consider a diffusion process $\{x(t)\}_{t=0}^{T}$ with $x(t) \in \mathbb{R}^n$, where $t \in [0,T]$ is a continuous time variable and $n$ denotes the image dimension. By choosing $x(0) \sim p_0$ and $x(T) \sim p_T$, $p_0$ to be the data distribution and $p_T$ to be the prior distribution, the diffusion process can be modeled as the solution of the following SDE:

$$dx = f(x,t)dt + g(t)dw \quad (5)$$

where $f \in \mathbb{R}^n$ and $g \in \mathbb{R}$ is the drift coefficient and the diffusion coefficient of $x(t)$, respectively. $w \in \mathbb{R}^n$ induces Brownian motion.

According to the reversibility of SDE, the reverse process of Eq. (5) can be expressed as another stochastic process:

$$dx = [f(x,t) - g(t)^2 \nabla_x \log p_t(x)]dt + g(t)d\bar{w} \quad (6)$$

where $dt$ is the infinitesimal negative time step, and $\bar{w}$ is a standard Wiener process for time flowing in reverse. The score term $\nabla_x \log p_t(x)$ can be approximated by a learned time-dependent score model $s_\theta(x_t,t)$. The SDE is then solved by means of some solver procedure, providing the basis for score-based generative modeling with SDEs.

### III. METHOD

#### A. Motivation of LR-KGM

Despite promising results, generative models still suffer from two major deficiencies: low data density regions and the manifold hypothesis [45], [46]. Inspired by [39], [47], we find that increasing the dimensionality of objects processed by generative models can effectively solve the above problems. In addition, from the perspective of machine learning, learning high-dimensional information is beneficial for generating representations [47]. More importantly, learning information in higher dimensions can further increase the number of input channels and realize the diversity of the generation. The higher the dimensionality, the greater the potential for generating diversity. On the other hand, we observe that the missing data can be better recovered by using the redundant information of Hankel matrix [48]. Given the above two aspects, a high-dimensional low-rank tensor assisted generative learning framework is proposed to enhance the network stability and performance.

In this work, LR-KGM handles multi-channel data through data structuring technology. Then, the processed data is mapped to high-dimensional tensor as the network input. As shown in Fig. 2, we first construct a large Hankel matrix from multi-channel MRI data. Afterward, the high-dimensional tensor is obtained by unfolding the Hankel matrix and applied to the diffusion model for training. By reverse-time SDE, we can smoothly mold random noise into data for sample generation in the reconstruction phase. We utilize the predictor-corrector (PC) sampling for the samples update step. At each time step, the numerical SDE solver gives an estimate of the sample. Langevin dynamics corrects the marginal distribution of the estimated sample, playing the role of a "corrector". Afterward, due to low-rank constraints on tensors, we further combine low-rank tensor iterations and generative iterations to achieve high-quality reconstruction. In short, the proposed LR-KGM involves two characteristics: (i) Learning the data prior information in higher-dimensional space, rather than the original space during training; (ii) Combining low-rank tensor iterations and generation iterations in iterative recovery in testing phase.

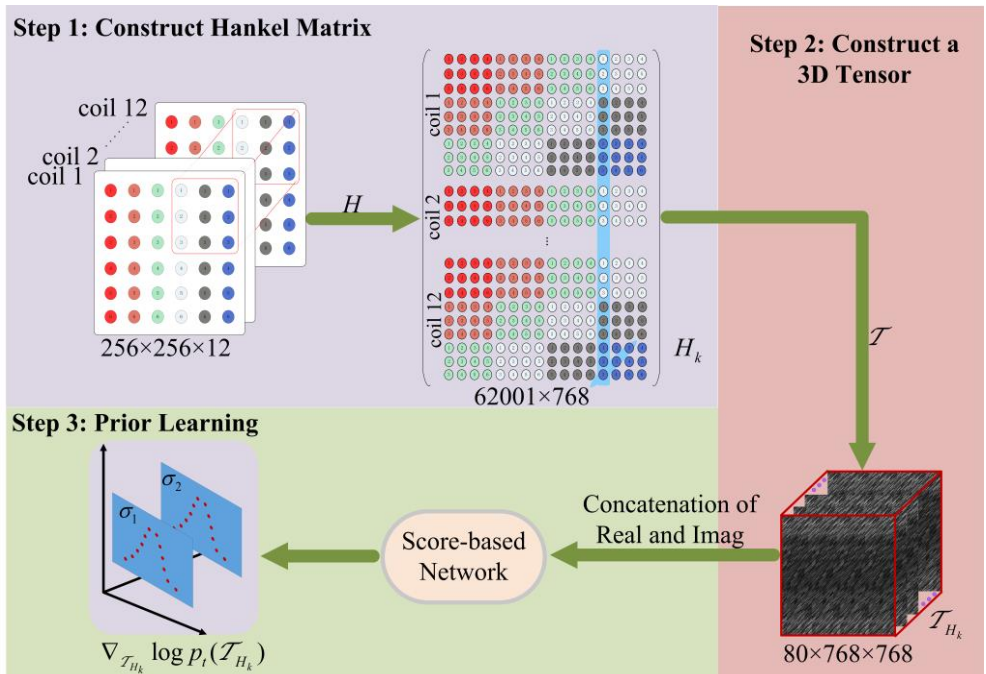

**Fig. 2.** The training flow chart of LR-KGM. The training process mainly consists of three steps. First, we construct a large Hankel matrix from k-space data. After that, we construct patches from Hankel matrix into a 3D tensor. Finally, we feed a 3D tensor to the network to capture the internal distribution.

## B. Low-rank Process

***Forming Hankel-induced tensor object.*** K-space data is constructed as a huge Hankel matrix, and large amounts of patches containing similar information can be obtained from the matrix. To generate diversity and facilitate better estimation of gradient information, a tensor is constructed through patches. In this work, 12 coils data is structured into a block-wise Hankel matrix of size $62001 \times 768$, of which columns are vectorized blocks from sliding kernels across the entire k-space. The size of the sliding kernel is $8 \times 8$. Then, we extract 80 patches of size $768 \times 768$ from it and stack them along a third dimension to form a third-order tensor for training. Fig. 3 demonstrates the process of Hankel matrix to low-rank tensor.

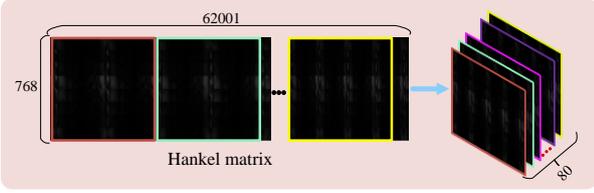

**Fig. 3.** The process of Hankel matrix to high-dimensional tensor.

***Low-rank rotation strategy.*** There are many traditional ways to deal with tensors. For example, one approach is to use the Tucker model [49] for tensor factorization. Another approach is to use the parallel factor analysis (Parafac) model [50], etc. For interested readers, please refer to [51].

To better handle high-dimensional tensors and excavate the correlations in the spatial dimension, we introduce the low-rank rotation strategy to impose low-rank constraint on tensors. Consider the tensor as multiple matrices and force the unfolding matrix along each mode of the tensor to be low rank. More specifically, we expand the tensor into a matrix along mode-i, applying SVD to process the matrix, and then fold it into a tensor. Later, the above process is repeated, the only difference being the modality of the tensor expansion. Since we construct a three-dimensional tensor, only three different directions of unrolling need to be done. The process can be expressed as follows:

$$\underset{\mathcal{A}, \mathbf{B}_n}{Min} \sum_{n=1}^{N} \alpha_n \|\mathbf{B}_n\|_* \quad s.t. \|\mathbf{A}^{(n)} - \mathbf{B}_n\|_F^2 \leq \tau, n = 1, 2, \cdots, N \quad (7)$$

where $\mathbf{B}_n$ are additional matrices. $\tau$ is a singular value thresholding that could be defined by the user. The matrix $\mathbf{A}^{(n)} \in \mathbb{R}^{L_n \times (L_1 \cdots L_{n-1} L_n L_{n+1} \cdots L_N)}$ is composed by taking all the mode-i vectors of $\mathcal{A}$ as its columns and can also be naturally seen as the mode-i flattening or unfolding of the tensor $\mathcal{A}$. An illustration of the tensor matricization is shown in Fig. 4. The details of Eq (7) can be found in literature [51].

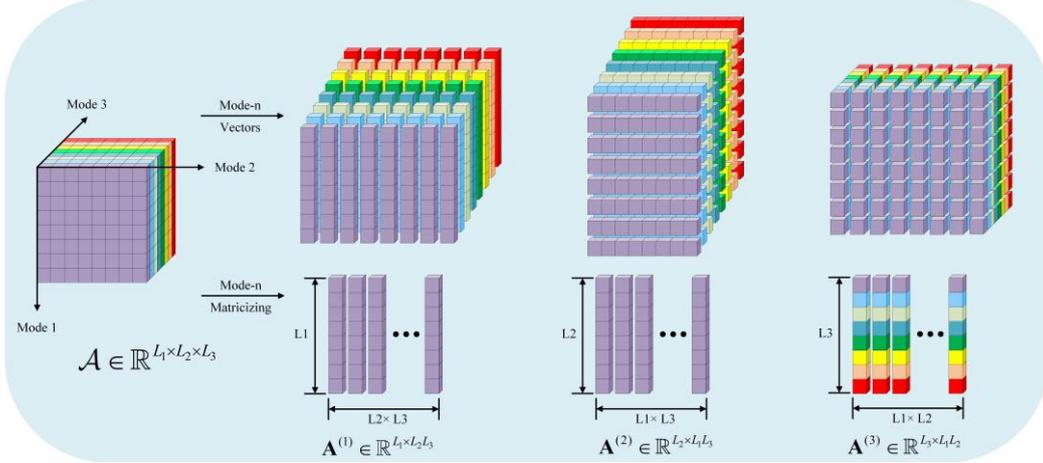

**Fig. 4.** Illustration of tensor matricization for a third-order tensor.

## C. Prior Learning in LR-KGM

In this study, the generative model learns the prior information in the high-dimensional space instead of the prior information in the original space. Thus, we can estimate the distribution of the corresponding data in tensor $\mathcal{T}_{H_k}$ via the scored-based network.

On the basis of Eq (5), the process of training can be described as:

$$d\mathcal{T}_{H_k} = f(\mathcal{T}_{H_k}, t)dt + g(\mathcal{T}_{H_k})dw \quad (8)$$

where $\mathcal{T}$ is the process of constructing the data into a 3D tensor.

According to *Song et al.* [38], we select to use Variance Exploding (VE) SDE for higher sample quality. $f = 0$ and $g = \sqrt{d[\sigma^2(t)]/dt}$ are applied to Eq. (8), resulting in

$$d\mathcal{T}_{H_k} = \sqrt{d[\sigma^2(t)]/dt} dw \quad (9)$$

where $\sigma(t)$ is a monotonically increasing function. The forward VE-SDE can be described into a Markov chain

$$\mathcal{T}_{H_k}^{i} = \mathcal{T}_{H_k}^{i-1} + \sqrt{\sigma_i^2 - \sigma_{i-1}^2} z_{i-1}, \ i = 1, \cdots, N \quad (10)$$

The score of a distribution can be estimated by performing score matching training on samples. Hence, we use denoising score matching to minimize Eq (11):

$$\theta^* = \arg\min_{\theta} \mathbb{E}_t \{\lambda(t) \mathbb{E}_{\mathcal{T}_{H_k}(0)} \mathbb{E}_{\mathcal{T}_{H_k}(t) | \mathcal{T}_{H_k}(0)} [ \\ \| s_\theta(\mathcal{T}_{H_k}(t), t) - \nabla_{\mathcal{T}_{H_k}(t)} \log p_{0t}(\mathcal{T}_{H_k}(t) | \mathcal{T}_{H_k}(0)) \|^2 ]\} \quad (11)$$

where $\lambda : [0,T] \longrightarrow \mathbb{R}_{>0}$ is a positive weighting function and $t$ is uniformly sampled over $[0,T]$. $\log p_{0t}(\mathcal{T}_{H_k}(t) | \mathcal{T}_{H_k}(0))$ is the transition kernel of Gaussian distribution.

## D. PI Reconstruction of LR-KGM

The reconstruction task in our work is divided into three subproblems that can be conducted alternatively, i.e., PC, the low-rank constraint on tensor (i.e.,

$\nabla_{\mathcal{T}_{H_k}} \log p_t(LR(k^g)^{i+1})$ ) and data consistency (DC) constraint on the measurement (i.e., $\nabla_{\mathcal{T}_{H_k}} \log p_t(Y | k^{i+1})$ ). The pipeline of the MRI reconstruction procedure in LR-KGM is shown in Fig. 5.

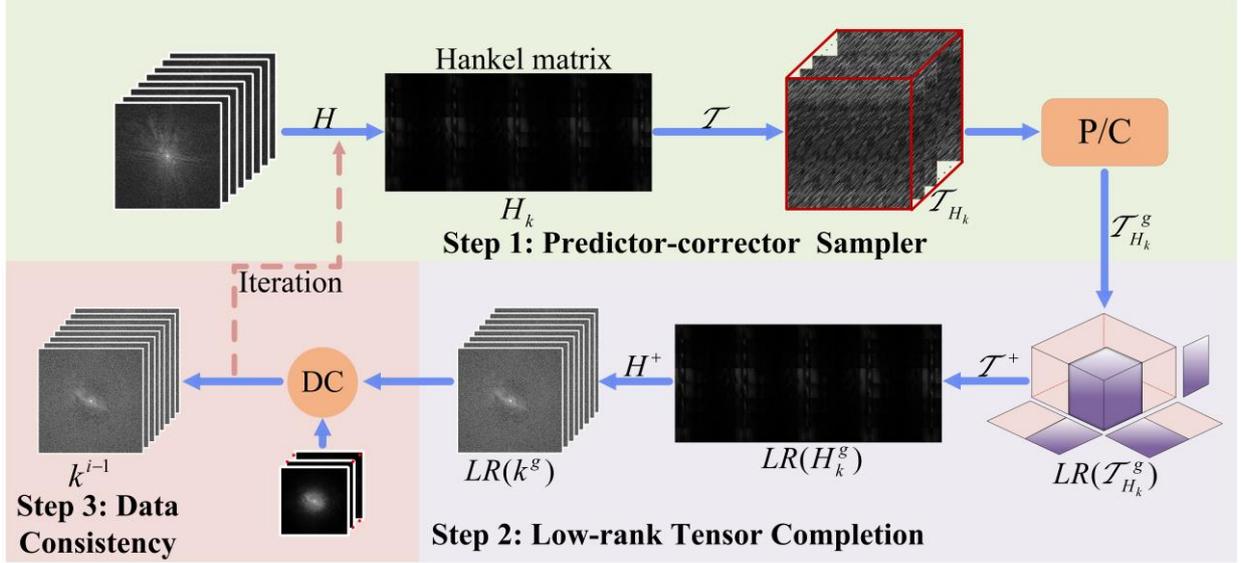

**Fig. 5.** Pipeline of the MRI reconstruction procedure of LR-KGM. The iterative reconstruction process mainly consists of three steps. Firstly, the noise data of k-space is constructed into a tensor and we use PC samplers to recover data in k-space from the trained network. Subsequently, low-rank rotation strategy is applied to recovered data. Finally, we perform data consistency and iterations on the k-space data.

*Step 1: PC sampler.* In the sampling process, a 12-channel k-space data is initialized from uniform noise as the input for the first iteration. Afterwards, these data are structured into high-dimensional tensor.

For samples update step, we utilize a PC framework to correct errors in the evolution of the discretized reverse-time SDE. The predictor is reverse diffusion SDE solver, which discretizes the reverse-time SDE in the same way as the forward one. It can be discretized as follows:

$$\begin{aligned}\mathcal{T}_{H_k}^{\ i} &= \mathcal{T}_{H_k}^{\ i+1} + (\sigma_{i+1}^2 - \sigma_i^2)\nabla_{\mathcal{T}_{H_k}} \log p_t(\mathcal{T}_{H_k}^{\ i+1}) + \sqrt{\sigma_{i+1}^2 - \sigma_i^2}\,z \\ &= \mathcal{T}_{H_k}^{\ i+1} + (\sigma_{i+1}^2 - \sigma_i^2)s_\theta(\mathcal{T}_{H_k}^{\ i+1}, \sigma_{i+1}) + \sqrt{\sigma_{i+1}^2 - \sigma_i^2}\,z \end{aligned} \quad (12)$$

where $z \sim N(0,1)$, $\mathcal{T}_{H_k}^{(0)} \sim p_0$, and we set $\sigma_0 = 0$ to simplify the notation. Langevin dynamics is considered as the corrector, which transforms any initial sample $\mathcal{T}_{H_k}^{(0)}$ to an approximate sample $p_t(\mathcal{T}_{H_k})$ with the following procedure:

$$\begin{aligned}\mathcal{T}_{H_k}^{\ i,j} &= \mathcal{T}_{H_k}^{\ i,j-1} + \varepsilon_i s_\theta(\mathcal{T}_{H_k}^{\ i,j-1}, \sigma_{i+1}) + \sqrt{2\varepsilon_i}\,z \\ j &= 1,2,\cdots,M; \quad i = N-1,\cdots,0 \end{aligned} \quad (13)$$

where $\varepsilon_i > 0$ is the step size.

In the circumstances of the low-rank constraint on tensor and DC, Eq. (12) can be described as conditional generation formulation:

$$\begin{aligned}\mathcal{T}_{H_k}^{\ i} &= \mathcal{T}_{H_k}^{\ i+1} + (\sigma_{i+1}^2 - \sigma_i^2)\nabla_{\mathcal{T}_{H_k}}[\log p_t(\mathcal{T}_{H_k}^{\ i+1}) \\ &\quad + \log p_t(LR(k^g)^{i+1}) + \log p_t(Y | k^{i+1})] + \sqrt{\sigma_{i+1}^2 - \sigma_i^2}\,z \\ &= \mathcal{T}_{H_k}^{\ i+1} + (\sigma_{i+1}^2 - \sigma_i^2)s_\theta(\mathcal{T}_{H_k}^{\ i+1}, \sigma_{i+1}) \\ &\quad + (\sigma_{i+1}^2 - \sigma_i^2)[\nabla_{\mathcal{T}_{H_k}} \log p_t(LR(k^g)^{i+1}) \\ &\quad + \nabla_{\mathcal{T}_{H_k}} \log p_t(Y | k^{i+1})] + \sqrt{\sigma_{i+1}^2 - \sigma_i^2}\,z \end{aligned} \quad (14)$$

where $Y$ is the under-sampled measure of the MRI reconstruction. The $\log p_t(\mathcal{T}_{H_k})$ is given by the prior model that represents information known beforehand about the true model parameter. The $\log p_t(LR(k^g))$ is derived from low-rank knowledge.

Similarly, Eq. (13) can be described as conditional generation formulation:

$$\begin{aligned}\mathcal{T}_{H_k}^{\ i,j} &= \mathcal{T}_{H_k}^{\ i,j-1} + \varepsilon_i \nabla_{\mathcal{T}_{H_k}}[\log p_t(\mathcal{T}_{H_k}^{\ i,j-1}) \\ &\quad + \log p_t(LR(k^g)^{i,j-1}) + \log p_t(Y | k^{i,j-1})] + \sqrt{2\varepsilon_i}\,z \end{aligned} \quad (15)$$

*Step 2: Low-rank tensor completion.* In low-rank tensor completion, the goal is to fill in missing entries of a partially known tensor and reduce data redundancy. After PC sampling, enforcing the low-rank constraint on tensor $\mathcal{T}_{H_k}^g$ yields tensor $LR(\mathcal{T}_{H_k}^g)$. Thus, low-rank constrained tensor is incorporated into the iterative process and combined with diffusion generation iterations to form a complete iterative process. Finally, $LR(k^g)$ is obtained from $LR(\mathcal{T}_{H_k}^g)$ by the corresponding inverse operations.

*Step 3: DC.* DC is simultaneously enforced in each iterative reconstruction step to ensure that the output is consistent with the original k-space information. From Eq. (15), we can obtain the subproblem regard to DC:

$$\underset{k}{Min}\{\|Mk - y\|_2^2 + \lambda \|k - LR(k^g)\|_2^2\} \quad (16)$$

The DC problem is solved via:

$$k(j) = \begin{cases} LR(k^g)(j), & if\ j \notin \Omega \\ \{\lambda[LR(k^g)(j)] + M^T y(j)\}/(1+\lambda), & if\ j \in \Omega \end{cases} \quad (17)$$

where $\Omega$ denotes an index set of the acquired k-space samples. $LR(k^g)(j)$ represents an entry at index $j$ in k-space. In the noiseless setting (i.e., $\lambda \to \infty$), we simply replace the $j$-th predicted coefficient by the original coefficient if it has been sampled. Then $k(j)$ go back to Step 1 for iterative reconstruction. After several iterations in k-space domain, the image is reconstructed by applying the inverse of Fouri-er encoding $I = F^{-1}k(j)$. The final reconstruction is obtained by combining the channels through SOS. It is worth noting that in Fig. 5, we only show a brief process of the reconstruction steps. In fact, the process starts with prediction and then make the correction after the execution of an iterative process. Algorithm 1 compactly represents the training and sampling process of LR-KGM.

---

**Algorithm 1: LR-KGM**

**Training stage**

**Dataset:** Several k-space data $k$

**1: Construct a 3D tensor** $\mathcal{T}_{H_k} = \mathcal{T}(H(k))$

  **Training** $s_\theta(\mathcal{T}_{H_k}, t) \simeq \nabla_{\mathcal{T}_{H_k}} \log p_t(\mathcal{T}_{H_k})$

**2: Output:** Trained $S_\theta(\mathcal{T}_{H_k}, t)$

**Reconstruction stage**

**Setting:** $N, M$

1: $\mathcal{T}_{H_k}^N \sim N(0, \sigma_T^2 \mathbf{I})$

2: For $i = N-1$ to $0$ do (Outer loop)

3:   $\mathcal{T}_{H_k}^i \leftarrow$ Predictor $(\mathcal{T}_{H_k}^{i+1}, \sigma_i, \sigma_{i+1})$

4:   $k^i = H^+(\mathcal{T}^+(LR(\mathcal{T}_{H_k}^i)))$

5:   Update Eq. (17) and $\mathcal{T}_{H_k}$

6:   For $j = 1$ to $M$ do (Inner loop)

7:     $\mathcal{T}_{H_k}^{i,j} \leftarrow$ Corrector$(\mathcal{T}_{H_k}^{i,j-1}, \sigma_i, \varepsilon_i)$

8:     $k^{i,j} = H^+(\mathcal{T}^+(LR(\mathcal{T}_{H_k}^{i,j})))$

9:     Update Eq. (17) and $\mathcal{T}_{H_k}$

10:  End for

11: End for

12: $k^{rec} = k^0$

13: Return $I_{sos} = \sqrt{\sum_{c=1}^{C} |(F^{-1}k_c^{rec})|^2}$

---

## IV. EXPERIMENTS

### A. Experimental Setup

**Datasets.** First, we use brain images from 12-channel *SIAT* dataset, which is provided by Shenzhen Institutes of Advanced Technology, the Chinese Academy of Science. Informed consents are obtained from the imaging subject in compliance with the institutional review board policy. The raw data is acquired from 3D turbo spin-echo (TSE) sequence with T2 weighting by a 3.0T whole-body MR system (SIEMENS MAGNETOM Trio Tim), which has 192 slices per slab, and the thickness of each slice is $0.86\ mm$. Typically, the field of view and voxel size are set to be $220 \times 220\ mm^2$ and $0.9 \times 0.9 \times 0.9\ mm^3$, respectively. The relevant imaging parameters encompass the size of image acquisition matrix is $256 \times 256$, echo time (TE) is $149\ ms$, repetition time (TR) is $2500\ ms$. At the training stage, we select 500 samples from *STAT* dataset with 12-channel for prior learning. In the meanwhile, we select 20 samples as experimental test dataset.

Aside from *SIAT* dataset, 12-channel *T2 Transversal brain* MR images with size of $256 \times 256$ are acquired with 3.0 T Siemens, whose FOV is $220 \times 220\ mm^2$ and TR/TE is $5000/91\ ms$. These brain images are used as test data in this article.

**Parameter settings.** In this subsection, parameter settings in the proposed algorithm are discussed. The batch size is set to 2 and Adam optimizer with $\beta_1=0.9$ and $\beta_2=0.999$ is utilized to optimize the network. For noise variance schedule, we fix $\sigma_{max} = 1$, $\sigma_{min} = 0.01$ and $r = 0.075$. For all the algorithms, we use $N = 1000$, $M = 1$ iterations. For the other parameters, we follow the settings in the work of Song *et al.* [23]. The training and testing experiments are performed with Operator Discretization Library (ODL) and PyTorch on a personal workstation with a GPU card (Tesla V100-PCIE-16GB).

**Evaluation metrics.** To quantitatively evaluate the performance of the various reconstruction models, we choose two classic metrics, namely peak signal to noise ratio (PSNR), and structural similarity index measure (SSIM). The PSNR describes the relationship of the maximum possible power of a signal with the power of noise corruption, and the SSIM is used to measure the similarity between the original images and reconstructed images. For the convenience of reproducibility, the source code and representative results are available at: *https://github.com/yqx7150/LR-KGM.*

### B. Reconstruction Comparisons

To evaluate the reconstruction performance, we compare the proposed method LR-KGM with other state-of-the-art algorithms, including ESPIRiT, LINDBERG, P-LORAKS, and SAKE. Moreover, we conduct the experiments under various sampling patterns (e.g., 2D Poisson, 2D Random sampling and 2D Partial sampling) and different acceleration factors (i.e., 4×, 6×, 8× and 10×).

**Test on T2 Transversal brain.** Fig. 6 depicts the qualitative reconstruction results of ESPIRiT, LINDBERG, P-LORAKS, SAKE, and LR-KGM along with their corresponding 5× magnified residual images. We conduct the experiments on *T2 Transversal brain* image under 2D Poisson sampling patterns with acceleration factor *R*=10. From Fig. 6, we can observe that there are significant residual artifacts and amplified noise that exist in the results obtained by ESPIRiT, LINDBERG, and P-LORAKS. Compared with ESPIRiT, LINDBERG, and P-LORAKS, the reconstructed error map of SAKE has less noise and clearer contour information. Nevertheless, the reconstructed image of LR-KGM preserves more small textural details and has less noise relative to the reference image. In LR-KGM, the overall reconstruction error is better than SAKE. Besides of the visual quality comparison, the quantitative is also highlighted. As can be seen in Table I, the reconstructions quality of different methods varies. Table I presents the quantitative comparisons of these methods at acceleration factors of 4 and 10. The image reconstructed by LR-KGM has larger PSNR and SSIM values than the other reconstructions under different acceleration factors, indicating the effectiveness of the proposed method. Specifically, the PSNR and SSIM of LR-KGM are the maximum values under the condition of *R*=4, which are 34.72 dB and 0.906, respectively.

**Test on SIAT brain.** Fig. 7 intuitively illustrates the representative reconstructions of the proposed method and competitive methods. The experimental results are obtained on 12-channel brain images from *SIAT* dataset under 2D Random sampling pattern with acceleration factor *R*=6. In line with previous results, LR-KGM provides better image quality with noise-like artifacts effectively suppressed com-pared with competitive results. Table II summarizes the quantitative PSNR/SSIM results with 2D Random sampling pattern and various sampling rates. Intuitively, the reconstruction results of LR-KGM outperform the comparison results, with high values of PSNR and SSIM.

TABLE I
PSNR AND SSIM COMPARISON WITH STATE-OF-THE-ART METHODS UNDER 2D POISSON SAMPLING PATTERNS WITH VARYING ACCELERATE FACTORS.

| *T2 Transversal Brain* | ESPIRIT | LINDBERG | P-LORAKS | SAKE | LR-KGM |
|---|---|---|---|---|---|
| 2D Poisson $R$=4 | 31.74/0.819 | 32.87/0.901 | 31.44/0.844 | 33.91/0.896 | **34.72/0.906** |
| 2D Poisson $R$=10 | 28.95/0.798 | 26.17/0.822 | 28.96/0.761 | 29.75/**0.823** | **31.10**/0.822 |

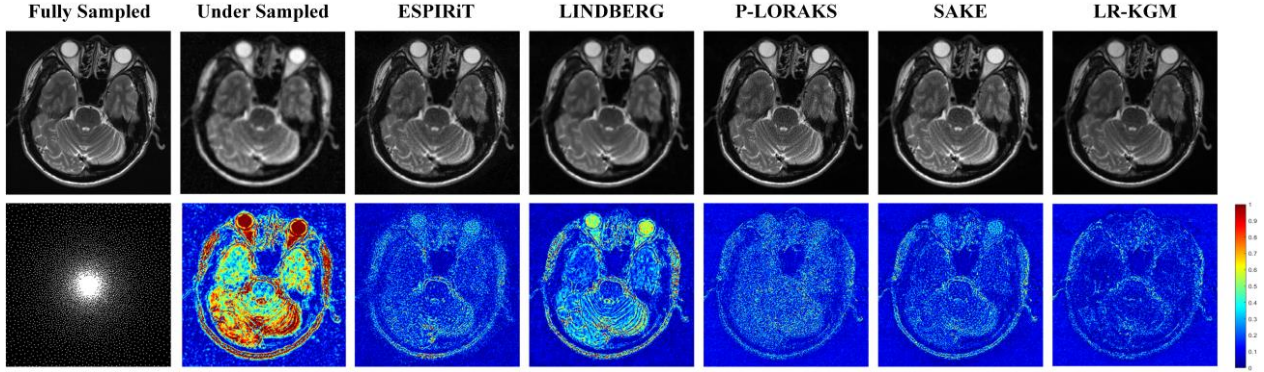

**Fig. 6.** Complex-valued PI reconstruction results at $R$=10 using 2D Poisson sampling with 12 coils. From left to right: Fully sampled, under-sampled, reconstruction by ESPIRiT, LINDBERG, P-LORAKS, SAKE, and LR-KGM. The intensity of residual maps is five times magnified.

TABLE II
PSNR AND SSIM COMPARISON WITH STATE-OF-THE-ART METHODS UNDER 2D RANDOM SAMPLING PATTERN WITH VARYING ACCELERATE FACTORS.

| *SIAT* | ESPIRIT | LINDBERG | P-LORAKS | SAKE | LR-KGM |
|---|---|---|---|---|---|
| 2D Random $R$=4 | 29.11/0.790 | 29.60/0.830 | 30.27/0.846 | 31.22/0.844 | **32.23/0.888** |
| 2D Random $R$=6 | 28.88/0.762 | 27.42/0.804 | 28.76/0.805 | 30.22/0.819 | **30.59/0.846** |

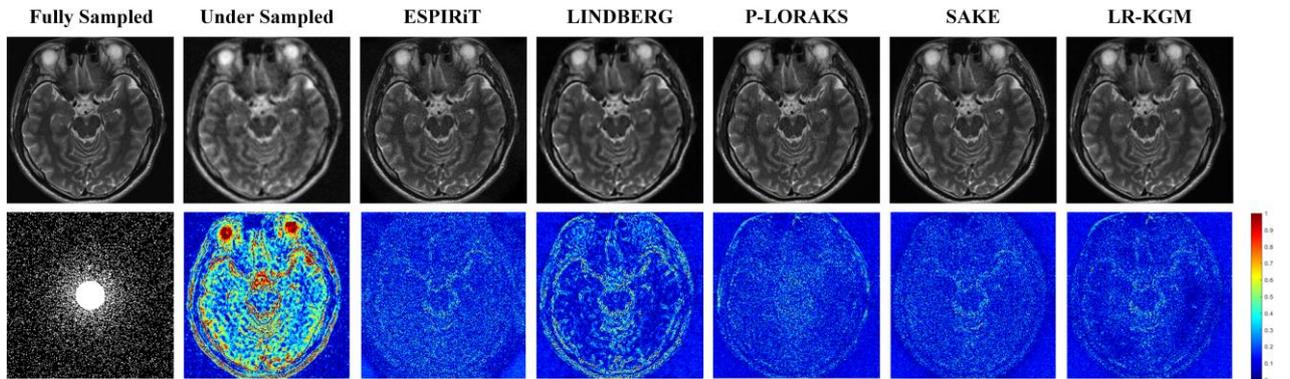

**Fig.7.** PI results by ESPIRiT, LINDBERG, P-LORAKS, SAKE and LR-KGM on *SIAT* at $R$=6 using 2D Random sampling mask.

Table III tabulates the numerical results of the proposed method and other comparison methods in terms of PSNR and SSIM metrics on *SIAT* brain images. In most cases, both of the test indicators of LR-KGM are higher than P-LORAKS and ESPIRiT, followed by SAKE. Notably, the average PSNR value of the results obtained by LR-KGM is nearly 2 dB higher than that in SAKE under 2D Random sampling pattern. Taking the case of $R$=4, the highest PSNR and SSIM values achieved by LR-KGM are 34.48 dB and 0.913, which are higher than the values of 32.37 dB and 0.866 obtained by SAKE. It can also be seen that LR-KGM can generate superior performance, even at high acceleration factors. These experiments demonstrate that the proposed model can be well generalized to various sampling patterns with different acceleration factors. Fig. 8 depicts the qualitative results of LR-KGM and competition methods, it can be seen that LR-KGM has much lower error with finer reconstruction details and much more edge information compared to other methods. In general, LR-KGM produces visually more convincing and accurate reconstruction with higher PSNR and SSIM values.

TABLE III
PSNR AND SSIM COMPARISON WITH STATE-OF-THE-ART METHODS UNDER DIFFERENT SAMPLING PATTERNS WITH VARYING ACCELERATE FACTORS.

| *SIAT* | ESPIRIT | P-LORAKS | SAKE | LR-KGM |
|---|---|---|---|---|
| 2D Poisson $R$=4 | 33.69/0.841 | 31.29/0.881 | 34.89/0.900 | **35.32/0.921** |
| 2D Poisson $R$=8 | 30.82/0.793 | 28.09/0.802 | 30.71/0.835 | **31.56/0.839** |
| 2D Random $R$=4 | 31.39/0.795 | 30.49/0.843 | 32.37/0.866 | **34.48/0.913** |
| 2D Random $R$=6 | 29.60/0.776 | 28.76/0.801 | 30.35/0.839 | **32.57/0.880** |

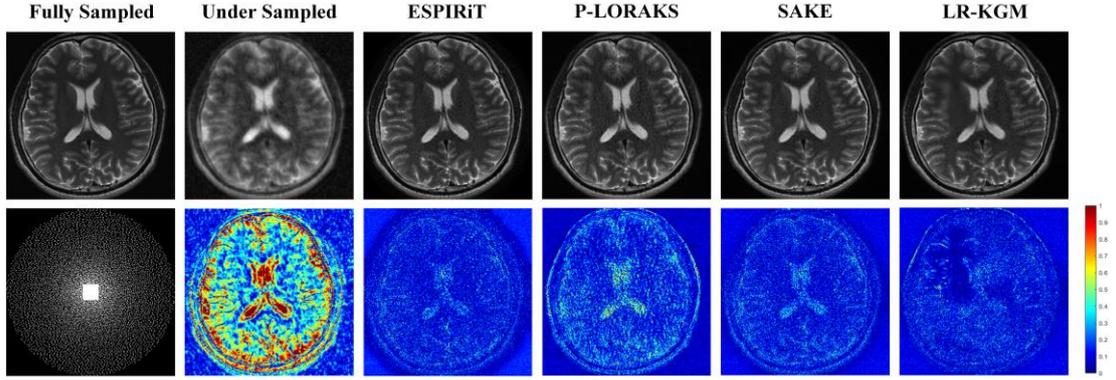

**Fig. 8.** PI reconstruction results by ESPIRiT, LINDBERG, P-LORAKS, SAKE and LR-KGM on *SIAT* image with 12 coils at *R*=8 using 2D Poisson sampling mask. The intensity of residual maps is five times magnified.

TABLE IV
PSNR AND SSIM COMPARISON UNDER 2D PARTIAL SAMPLING PATTERNS WITH VARYING ACCELERATE FACTORS.

| *SIAT* | Zero-filled | SAKE | LR-KGM |
|---|---|---|---|
| 2D Partial *R*=3 | 22.57/0.767 | 29.76/0.927 | **30.06/0.932** |
| 2D Partial *R*=6 | 20.14/0.651 | 28.74/0.862 | **29.46/0.873** |

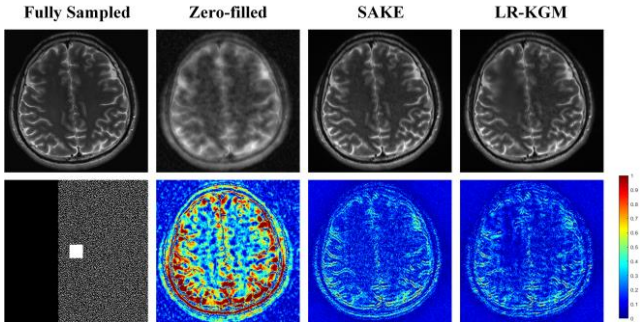

**Fig. 9.** PI reconstruction results by Zero-filled, SAKE and LR-KGM on *SIAT* image with 12 coils at *R*=6 Partial sampling mask. The intensity of residual maps is five times magnified.

Table IV lists the quantitative comparisons of the proposed LR-KGM with Zero-filled and SAKE, respectively. The experimental results are obtained on 12-channel *SIAT* brain images under Partial sampling mask with acceleration factors *R*=3 and *R*=6. Compared to method SAKE, the reconstruction performance of LR-KGM improves with higher PSNR and SSIM values under two different types of acceleration factors. Specifically, compared with SAKE, the PSNR value of LR-KGM reconstructed images increases by about 1dB. Based on the residual maps shown in Fig. 9, it can be concluded that LR-KGM is superior to SAKE in terms of reconstruction quality and structure details preservation. The above experimental results show that the model can be well generalized to various sampling modes with different acceleration factors.

### C. Convergence Analysis

In this section, the training loss curve and iterative convergence curves of LR-KGM are described. First, Fig. 10 shows how the training loss varies with the number of iterations steps. It can be observed that as the number of iterations steps increases, the curve of the training loss gradually converges to a low point. Although there are fluctuations in the iterations process, the overall network development trend gradually converges. In particular, we can easily see from the training loss curve that the proposed network converged at approximately the 20000-th iteration.

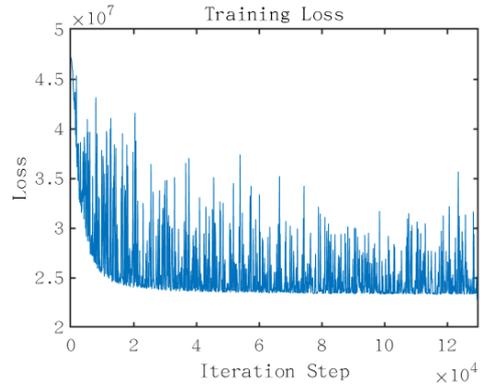

**Fig. 10.** The training loss curve of LR-KGM.

Second, we experimentally investigate the convergence of LR-KGM and SAKE with the number of iterations. We randomly select an example of reconstructed image from *SIAT* dataset using 2D Random sampling pattern with acceleration factor *R*=6. As shown in Fig. 11, the PSNR curve of LR-KGM converges faster. In addition, when the number of iterations is small, the PSNR curves of both LR-KGM and SAKE rise rapidly as the number of iterations increases. But when the number of iterations reaches a certain value, the PSNR values of SAKE will decrease with the increasing number of iterations while LR-KGM maintains a stable trend. More importantly, the PSNR of LR-KGM is always higher than that of SAKE. In Fig. 12, the SSIM curve of LR-KGM rises faster in early iterations and first converges to a stable point. However, the SSIM of SAKE drops sharply, while that of LR-KGM remains stable as the increasing.

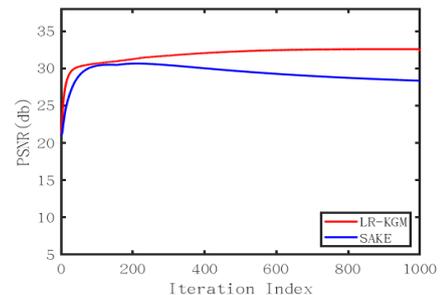

**Fig. 11.** Convergence curves of SAKE and LR-KGM in terms of PSNR versus the iterations number when reconstructing the brain image from *SIAT* using 2D Random sampling pattern with acceleration factor *R*=6.

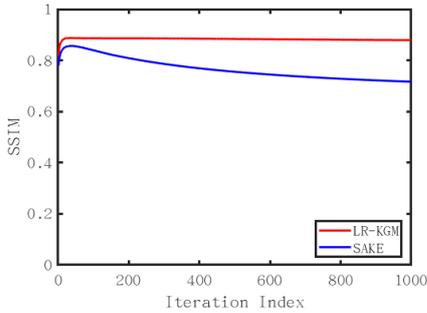

**Fig. 12**. Convergence curves of SAKE and LR-KGM in terms of SSIM versus the iterations number when reconstructing the brain image from *SIAT* using 2D Random sampling pattern with acceleration factor *R*=6.

### D. Ablation Study

We select different number of training data samples as the network input to verify the generation capability of LR-KGM, and two quantitative metrics are listed in Table V. Under the Random sampling pattern, PSNR and SSIM reach the maximum values with 500 images as input. However, the number of samples does not lead to a large difference in the reconstruction effect. For example, the PSNR for training one sample is only about 0.2 dB different from the PSNR for training 200 samples. Therefore, reducing the number of training samples to achieve a more outstanding reconstruction effect is what we will pursue in future work.

TABLE V
PSNR AND SSIM COMPARISON WITH DIFFERENT NUMBER OF INPUT IMAGES UNDER 2D RANDOM WITH VARYING ACCELERATE FACTORS.

| *SIAT* | 1 | 200 | 500 |
|---|---|---|---|
| 2D Random *R*=4 | 34.41/0.910 | 34.56/0.914 | **34.60/0.915** |
| 2D Random *R*=6 | 32.36/0.875 | 32.56/0.883 | **32.63/0.884** |

Then, we verify the effect of singular value thresholding on the local error and overall error of the reconstructed image. In Fig. 13, we can observe that the reconstruction results have larger local errors when the singular value thresholding is small. Nonetheless, the overall reconstruction effect is better. On the contrary, the reconstruction result will not have large local error when the singular value thresholding is high. However, the overall reconstruction effect is poor. Table VI also quantitatively proves this phenomenon.

TABLE VI
PSNR AND SSIM COMPARISON OF DIFFERENT SINGULAR VALUE THRESHOLDING SIZES UNDER DIFFERENT SAMPLING PATTERNS WITH VARYING ACCELERATE FACTORS.

| *STAT* | $\tau$ =48 | $\tau$ =56 | $\tau$ =64 | $\tau$ =72 | $\tau$ =80 |
|---|---|---|---|---|---|
| 2D Poisson *R*=4 | 33.46/0.914 | 33.90/**0.915** | **34.13**/0.914 | 33.55/0.902 | 33.35/0.897 |
| 2D random *R*=6 | 30.67/0.866 | 31.38/0.870 | **31.85/0.871** | 31.60/0.860 | 31.14/0.840 |

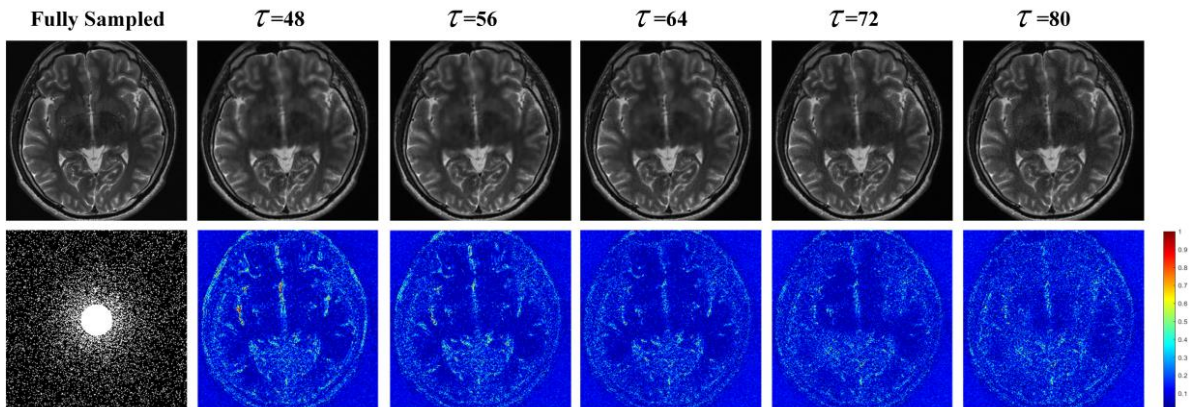

**Fig. 13.** Reconstruction results of models trained using different singular value thresholding size choices on *SIAT* dataset. From left to right: $\tau$ =48 model, $\tau$ =56 model, $\tau$ =64 model, $\tau$ =72 model and $\tau$ =80 model.

## V. CONCLUSION

This work designed a low-rank tensor assisted k-space generative model LR-KGM for parallel imaging reconstruction. The proposed LR-KGM performed generative learning in a high-dimensional space, which increases the dimensionality of the processed object (i.e., the number of input channels), and the possibility of generative diversity increases accordingly. At the iterative stage, low-rank constraint was imposed on tensor, and generation iterations and high-dimensional low-rank iterations were alternately performed. Experiment results verified that LR-KGM can produce better reconstruction performance under different sampling patterns with large acceleration factors, keeping higher PSNR and SSIM values compared to state-of-the-arts.

## REFERENCE

[1] M. Lustig, D. L. Donoho, J. M. Santos, *et al.*, "Compressed sensing MRI," *IEEE Signal Process. Mag.*, vol. 25, no. 2, pp. 72-82, 2008.
[2] M. Akcakaya, S. Nam, P. Hu, M. H. Moghari, L. H. Ngo, V. Tarokh, W. J. Manning, and R. Nezafat, "Compressed sensing with wavelet domain dependencies for coronary MRI: a retrospective study," *IEEE Trans. Med. Imaging,* vol. 30, no. 5, pp. 1090-1099, 2011.
[3] K. T. Block, M. Uecker, and J. Frahm, "Undersampled radial MRI with multiple coils. Iterative image reconstruction using a total variation constraint," *Magn. Reson. Med.,* vol. 57, no. 6, pp. 1086-1098, 2007.
[4] H. Jung, K. Sung, K. S. Nayak, E. Y. Kim, and J. C. Ye, "k-t FOCUSS: a general compressed sensing framework for high resolution dynamic MRI," *Magn. Reson. Med.,* vol. 61, no. 1, pp. 103-116, 2009.
[5] D. K. Sodickson and W. J. Manning, "Simultaneous acquisition of spatial harmonics (SMASH): Fast imaging with radiofrequency coil arrays," *Magn. Reson. Med.,* vol. 38, no. 4, pp. 591-603, 1997.
[6] K. P. Pruessmann, M. Weiger, M. B. Scheidegger, and P. Boesiger, "SENSE: Sensitivity encoding for fast MRI," *Magn. Reson. Med.,* vol.

[7] M. A. Griswold, P. M. Jakob, R. M. Heidemann, M. Nittka, V. Jellus, J. Wang, B. Kiefer, and A. Haase, "Generalized autocalibrating partially parallel acquisitions (GRAPPA)," *Magn. Reson. Med.,* vol. 47, no. 6, pp. 1202-1210, 2002.

[8] S. G. Lingala, Y. Hu, E. DiBella, and M. Jacob, "Accelerated dynamic MRI exploiting sparsity and low-rank structure: k-t SLR," *IEEE Trans. Med. Imaging*, vol. 30, no. 5, pp. 1042-1054, 2011.

[9] R. Otazo, E. Candes, and D. K. Sodickson, "Low-rank plus sparse matrix decomposition for accelerated dynamic MRI with separation of background and dynamic components," *Magn. Reson. Med*, vol. 73, no. 3, pp. 1125-1136, 2015.

[10] X. Qu, M. Mayzel, Z. Chen, and V. Orekhov, "Accelerated NMR spectroscopy with low-rank reconstruction," *Angew. Chem.,* vol. 127, no. 3, pp. 866-868, 2015.

[11] X. Zhang, D. Guo, Y. Huang, *et al.,* "Image reconstruction with low-rankness and self-consistency of k-space data in parallel MRI," *Medical Image Analysis*, 2020.

[12] J. P. Haldar and J. Zhuo, "P-LORAKS: low-rank modeling of local k-space neighborhoods with parallel imaging data," *Magn. Reason. Med.*, vol. 75, no. 4, pp. 1499-1514, 2016.

[13] X. Chen, W. Wu, and M. Chiew, "Locally structured low-rank MR image reconstruction using submatrix constraints," *IEEE 19th International Symposium on Biomedical Imaging (ISBI)*, pp. 1-4, 2022.

[14] P. J. Shin, P. E. Larson, M. A. Ohliger, M. Elad, and M. Lustig, *et al.,* "Calibration-less parallel imaging reconstruction based on structured low-rank matrix completion," *Magn. Reson. Med.,* vol. 72, no. 4, pp. 959-970, 2014.

[15] J. P. Haldar, "Low-rank modeling of local k-space neighborhoods (LORAKS) for constrained MRI," *IEEE Trans. Med. Imaging*, vol. 33, no. 3, pp. 668-681, 2014.

[16] K. H. Jin, D. Lee, and J. C. Ye, "A general framework for compressed sensing and parallel MRI using annihilating filter based low-rank hankel matrix," *IEEE Trans. Compu. Imag.*, vol. 2, no. 4, pp. 480-495, 2016.

[17] J. He, Q. Liu, A. G. Christodoulou, C. Ma, and F. Lam, "Accelerated high-dimensional MR imaging with sparse sampling using low-rank tensors," *IEEE Trans. Med. Imag.,* vol. 35, no. 9, pp. 2119-2129, 2016.

[18] N. Kargas, S. Weingärtner, and N. D. Sidiropoulos, "Low-rank tensor regularization for improved dynamic quantitative magnetic resonance imaging," in *SPARS*, 2017.

[19] Y. Liu, T. Liu, and C. Zhu, "Smooth robust tensor principal component analysis for compressed sensing of dynamic MRI," *Pattern Recogn.*, vol. 102, 2020.

[20] Y. Yu, J. Jin, F. Liu, and S. Crozier, "Multidimensional compressed sensing MRI using tensor decomposition-based sparsifying transform," *PloS one*, vol. 9, no. 6, 2014.

[21] M. Mardani, L. Ying, and G. Giannakis, "Accelerating dynamic MRI via tensor subspace learning," in *Proc. ISMRM,* 2015.

[22] Q. Tian, Z. Li, Q. Fan, *et al.,* "SDnDTI: Self-supervised deep learning-based denoising for diffusion tensor MRI," *NeuroImage*, vol. 253, 2022.

[23] Y. Liu, Z. Yi, Y. Zhao, F. Chen, and Y. Feng, "Calibrationless parallel imaging reconstruction for multislice MR data using low-rank tensor completion," *Magn. Reson. Med.*, vol. 85, no. 2, pp. 897-911, 2021.

[24] Y. Zhao, Z. Yi, and L. Xiao, "Calibrationless multi-slice Cartesian MRI via orthogonally alternating phase encoding direction and joint low-rank tensor completion," *NMR in Biomedicine*, 2022.

[25] Y. LeCun, Y. Bengio, and G. Hinton, "Deep learning," *Nature*, vol. 521, pp. 436-444, May 2015.

[26] Q. Liu, Q. Yang, H. Cheng, S. Wang, M. Zhang, and D. Liang, "Highly undersampled magnetic resonance imaging reconstruction using autoencoding priors," *Magn. Reason. Med.*, vol. 83, no. 1, pp. 322-336, 2020.

[27] Y. Liu, Q. Liu, M. Zhang, Q. Yang, S. Wang, and D. Liang, "IFR-Net: Iterative feature refinement network for compressed sensing MRI," *IEEE Trans. Comput. Imag.*, vol. 6, pp. 434-446, 2019.

[28] Y. Yang, J. Sun, H. Li, and Z. Xu, "ADMM-Net: a deep learning approach for compressive sensing MRI," *arXiv:1705.06869*.

[29] Y. Arefeen, O. Beker, J. Cho, *et al.* "Scan-specific artifact reduction in k-space (SPARK) neural networks synergize with physics-based reconstruction to accelerate MRI," *Magn. Reson. Med.*, vol. 87, no. 2, pp. 764-780, 2022.

[30] M. Akçakaya, S. Moeller, S. Weingärtner, *et al.* "Scan-specific robust artificial-neural-networks for k-space interpolation (RAKI) reconstruction: Database-free deep learning for fast imaging," *Magn. Reson. Med.*, vol. 81, no. 1, pp. 439-453, 2019.

[31] A. Bora, A. Jalal, E. Price, and A.G. Dimakis, "Compressed sensing using generative models," *Int. Conf. on Mach. Learn.*, pp. 537-546, 2017.

[32] H. Chung and J. C. Yea, "Score-based diffusion models for accelerated MRI," *arXiv preprint arXiv:2110.05243*, 2021.

[33] A. Jalal, M. Arvinte, Daras G, E. Price, A. Dimakis, and J. Tamir, "Robust compressed sensing MRI with deep generative priors," *Adv. Neural Inf. Process. Syst.*, vol. 34, 2021.

[34] Y. Xie and Q. Li, "Measurement-conditioned denoising diffusion probabilistic model for under-sampled medical image reconstruction," *arXiv preprint arXiv:2203.03623*, 2022.

[35] G. Luo, N. Zhao, W. Jiang, E.S. Hui, and C. Peng, "MRI reconstruction using deep Bayesian estimation," *Magn. Reason. Med.,* vol. 84, no. 4, pp. 2246-2261, 2020.

[36] Y. Guan, Z. Tu, S. Wang, Q. Liu, Y. Wang, and D. Liang, "MRI reconstruction using deep energy-based model," *arXiv preprint arXiv:2109.03237*, 2021.

[37] Y. Song and S. Ermon, "Generative modeling by estimating gradients of the data distribution," *Proc. Adv. Neural Inf. Process. Syst.*, pp. 11895-11907, 2019.

[38] Y. Song, J. Sohl-Dickstein, D. P. Kingma, A. Kumar, S. Ermon, and B. Poole, "Score-based generative modeling through stochastic differential equations," *arXiv preprint arXiv:2011.13456,* 2020.

[39] C. Quan, J. Zhou, Y. Zhu, Y. Chen, S. Wang, D. Liang, and Q. Liu, "Homotopic gradients of generative density priors for MR image reconstruction," *IEEE Trans. Med. Imag.*, vol. 40, no. 12, pp. 3265-3278, 2021.

[40] Z. Zhou, Y. Guo, and Y. Wang, "Handheld ultrasound video high-quality reconstruction using a low-rank representation multipathway generative adversarial network," *IEEE Trans. Neural Netw. Learn. Syst.,* vol. 32, pp. 575-588, 2020.

[41] S. Li, J. Zhu, and C. Miao, "A generative word embedding model and its low rank positive semidefinite solution," *arXiv preprint arXiv:1508.03826*.

[42] S. Lloyd and C. Weedbrook, "Quantum generative adversarial learning," *Phys. Rev. Lett.,* vol. 121, 2018.

[43] Z. Ding, M. Shao, and Y. Fu, "Generative zero-shot learning via low-rank embedded semantic dictionary," *IEEE Trans. Pattern Anal. Mach. Intell.,* vol. 41, no. 12, pp. 2861-2874, 2018.

[44] J. Cocola, P. Hand, and V. Voroninski, "Nonasymptotic guarantees for spiked matrix recovery with generative priors," in *NIPS,* pp. 15185-15197, 2020.

[45] H. Narayanan and S. Mitter, "Sample complexity of testing the manifold hypothesis," in *Proc. Adv. Neural Inf. Process. Syst.,* pp. 1786-1794, 2010.

[46] S. Rifai, Y. Dauphin, P. Vincent, Y. Bengio, and X. Muller, "The manifold tangent classifier," in *Proc. Adv. Neural Inf. Process. Syst.,* pp. 2294-2302, 2011.

[47] K. Hong, C. Wu, C. Yang, M. Zhang, Y. Lu, Y. Wang and Q. Liu, "High-dimensional Assisted Generative Model for Color Image Restoration," *arXiv preprint arXiv:2108.06460*.

[48] P. J. Shin, P. E. Larson, M. A. Ohliger, *et al.,* "Calibrationless parallel imaging reconstruction based on structured low-rank matrix completion," *Magn. Reson. Med.*, vol. 72, no. 4, pp. 959-970, 2014.

[49] L.R. Tucker, "Some Mathematical Notes on Three-Mode Factor Analysis," *Psychometrika,* vol. 31, pp. 279-311, 1966.

[50] R.A. Harshman, "Foundations of the Parafac Procedure: Models and Conditions for an "Explanatory" Multi-Modal Factor Analysis," *UCLA Working Papers in Phonetics,* vol. 16, pp. 1-84, 1970.

[51] J. Liu, P. Musialski, P. Wonka and J. Ye, "Tensor completion for estimating missing values in visual data," *IEEE Trans. Pattern Anal. Machine Intell.*, vol. 35, pp. 208-220, 2012.